\documentclass[11pt]{article}
\usepackage{coling2020}
\usepackage{times}
\usepackage{url}
\usepackage{latexsym}
\usepackage{graphicx}
\usepackage{subcaption}
\usepackage{listings}
\usepackage{multirow}
\usepackage{amsmath}
\usepackage{textcomp}
\usepackage{booktabs}
\usepackage{pifont}
\usepackage[table,xcdraw]{xcolor}

\definecolor{titlecolor}{RGB}{0,191,255}
\definecolor{abstractcolor}{RGB}{255,182,193}
\definecolor{authorcolor}{RGB}{0,0,139}
\definecolor{sectioncolor}{RGB}{0,0,255}
\definecolor{footercolor}{RGB}{255,255,255}
\definecolor{equationcolor}{RGB}{255,0,0}
\definecolor{figurecolor}{RGB}{255,0,255}
\definecolor{captioncolor}{RGB}{0,255,0}
\definecolor{tablecolor}{RGB}{0,255,255}
\definecolor{paracolor}{RGB}{176,196,222}
\definecolor{listcolor}{RGB}{48, 32, 112}
\definecolor{refcolor}{RGB}{48, 144, 0}

\colingfinalcopy 

\title{DocBank: A Benchmark Dataset for Document Layout Analysis}

\author{Minghao Li$^{1*}$, Yiheng Xu$^{2}$\thanks{Equal contributions during internship at Microsoft Research Asia.}, Lei Cui$^{2}$, Shaohan Huang$^{2}$, \\
\textbf{Furu Wei$^{2}$, Zhoujun Li$^{1}$, Ming Zhou$^{2}$}\\
$^{1}$Beihang University\\
$^{2}$Microsoft Research Asia\\
\texttt{\{liminghao1630,lizj\}@buaa.edu.cn}\\
\texttt{\{v-yixu,lecu,shaohanh,fuwei,mingzhou\}@microsoft.com} \\
}

\begin{document}
\maketitle
\begin{abstract}
Document layout analysis usually relies on computer vision models to understand documents while ignoring textual information that is vital to capture. Meanwhile, high quality labeled datasets with both visual and textual information are still insufficient. In this paper, we present \textbf{DocBank}, a benchmark dataset that contains 500K document pages with fine-grained token-level annotations for document layout analysis. DocBank is constructed using a simple yet effective way with weak supervision from the \LaTeX{} documents available on the arXiv.com. With DocBank, models from different modalities can be compared fairly and multi-modal approaches will be further investigated and boost the performance of document layout analysis. We build several strong baselines and manually split train/dev/test sets for evaluation. Experiment results show that models trained on DocBank accurately recognize the layout information for a variety of documents. The DocBank dataset is publicly available at \url{https://github.com/doc-analysis/DocBank}.
\end{abstract}

\section{Introduction}

%
%
\blfootnote{
    
    This work is licensed under a Creative Commons 
    Attribution 4.0 International License.
    License details:
    \url{http://creativecommons.org/licenses/by/4.0/}.
}

Document layout analysis is an important task in many document understanding applications as it can transform semi-structured information into a structured representation, meanwhile extracting key information from the documents. It is a challenging problem due to the varying layouts and formats of the documents. Existing techniques have been proposed based on conventional rule-based or machine learning methods, where most of them fail to generalize well because they rely on hand crafted features that may be not robust to layout variations. Recently, the rapid development of deep learning in computer vision has significantly boosted the data-driven image-based approaches for document layout analysis. Although these approaches have been widely adopted and made significant progress, they usually leverage visual features while neglecting textual features from the documents. Therefore, it is inevitable to explore how to leverage the visual and textual information in a unified way for document layout analysis. 



Nowadays, the state-of-the-art computer vision and NLP models are often built upon the pre-trained models~\cite{peters2018deep,radford2018improving,bert,lample2019cross,yang2019xlnet,Dong2019UnifiedLM,2019t5,Xu2019LayoutLMPO} followed by fine-tuning on specific downstream tasks, which achieves very promising results. However, pre-trained models not only require large-scale unlabeled data for self-supervised learning, but also need high quality labeled data for task-specific fine-tuning to achieve good performance. For document layout analysis tasks, there have been some image-based document layout datasets, while most of them are built for computer vision approaches and they are difficult to apply to NLP methods. In addition, image-based datasets mainly include the page images and the bounding boxes of large semantic structures, which are not fine-grained token-level annotations. Moreover, it is also time-consuming and labor-intensive to produce human-labeled and fine-grained token-level text block arrangement. Therefore, it is vital to leverage weak supervision to obtain fine-grained labeled documents with minimum efforts, meanwhile making the data be easily applied to any NLP and computer vision approaches. 


\begin{figure*}[t]
\centering
    \begin{subfigure}[b]{0.235\textwidth}
        \includegraphics[width=\textwidth]{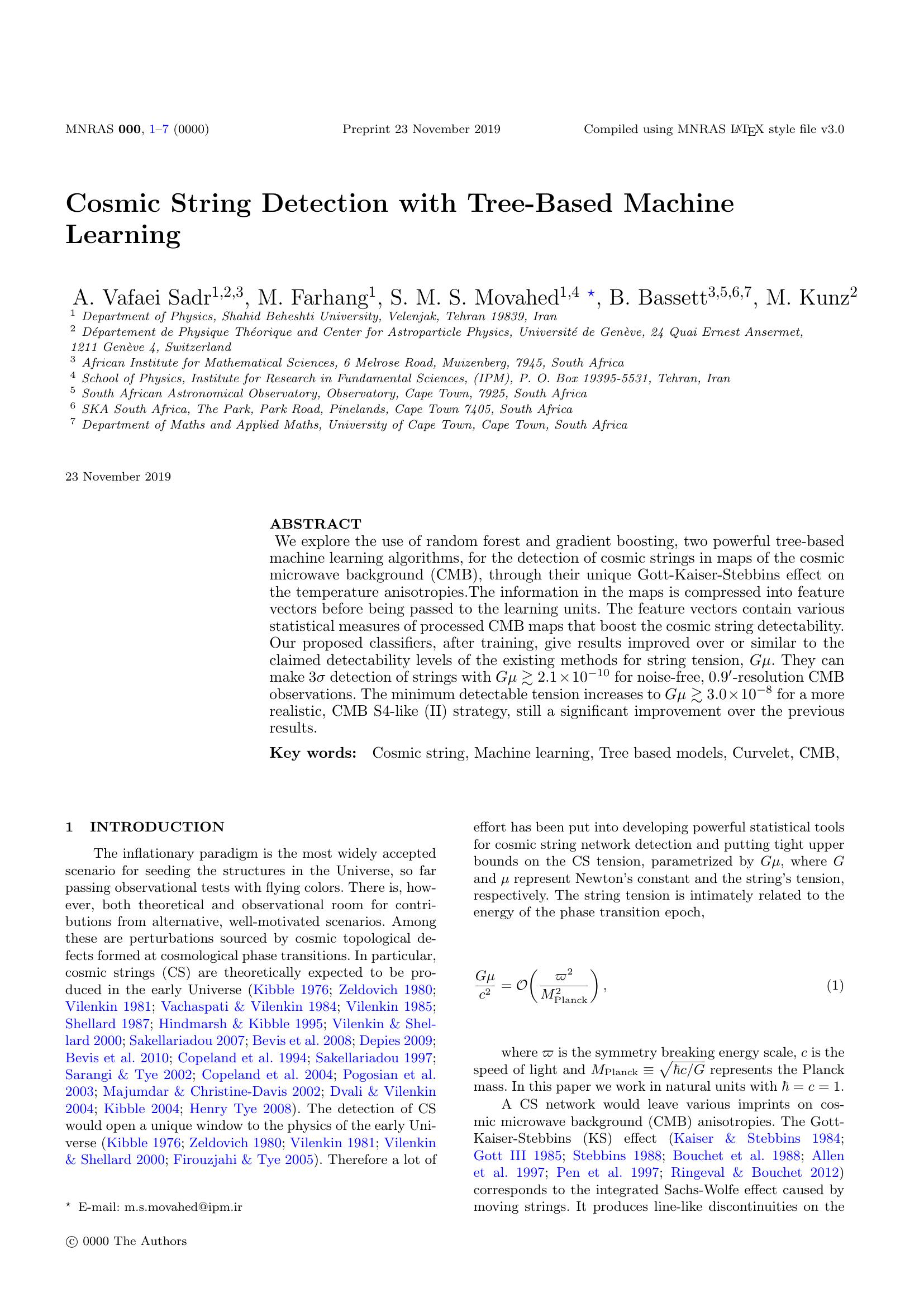}
        \caption{}
        \label{fig:3a}
    \end{subfigure}
    ~ 
    \begin{subfigure}[b]{0.235\textwidth}
        \includegraphics[width=\textwidth]{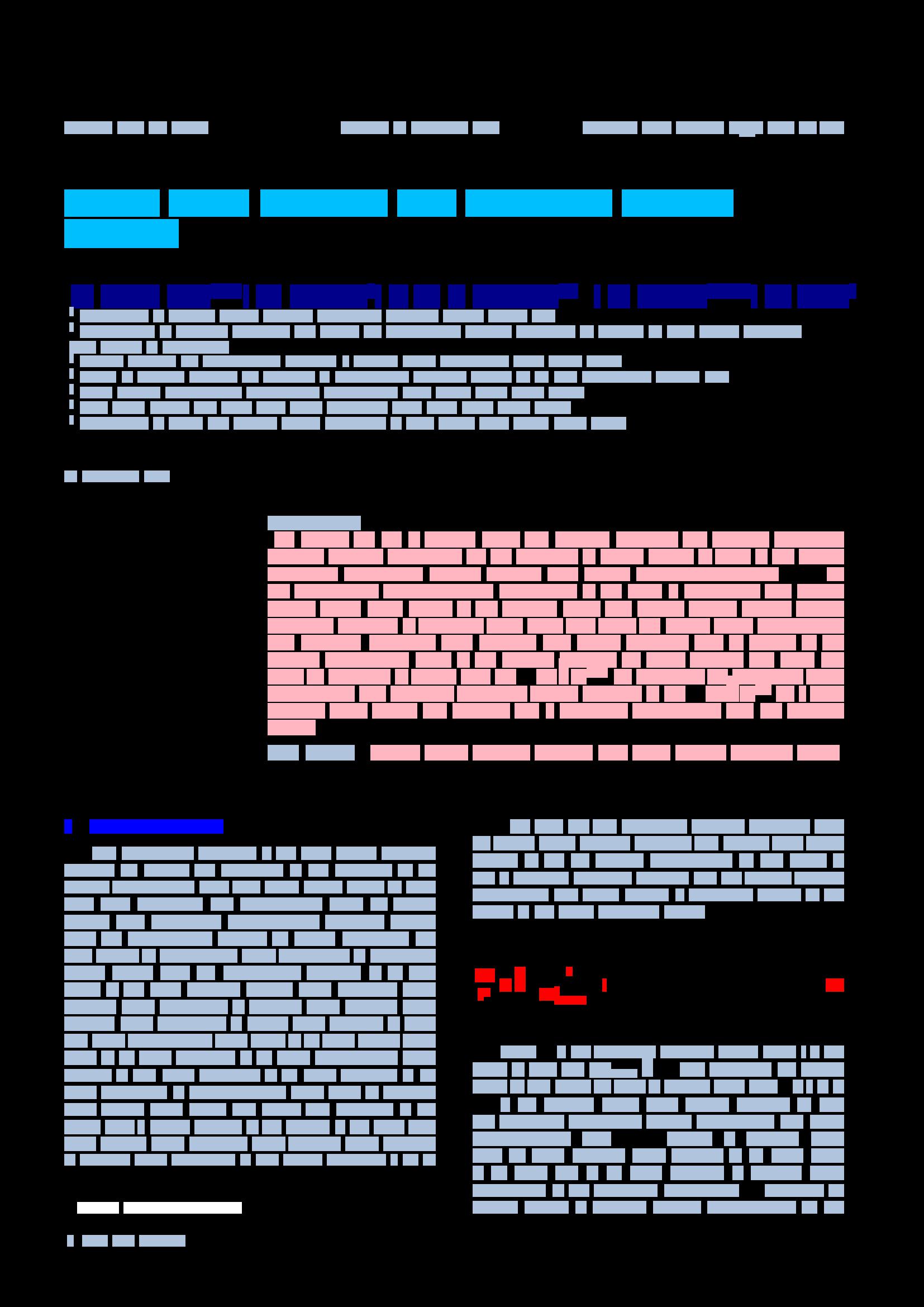}
        \caption{}
        \label{fig:3b}
    \end{subfigure}
    ~ 
    \begin{subfigure}[b]{0.235\textwidth}
        \includegraphics[width=\textwidth]{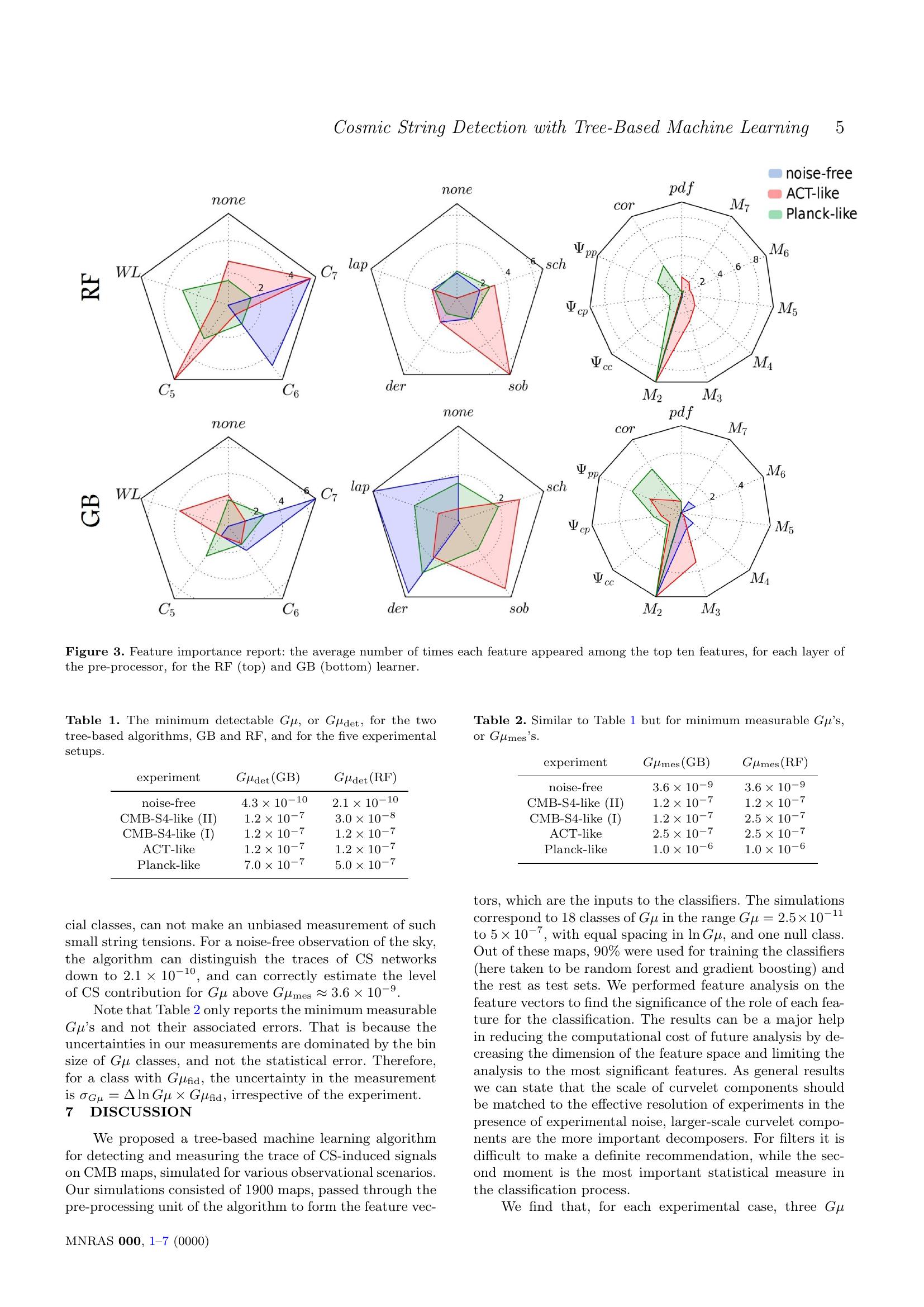}
        \caption{}
        \label{fig:3c}
    \end{subfigure}
    ~
    \begin{subfigure}[b]{0.235\textwidth}
        \includegraphics[width=\textwidth]{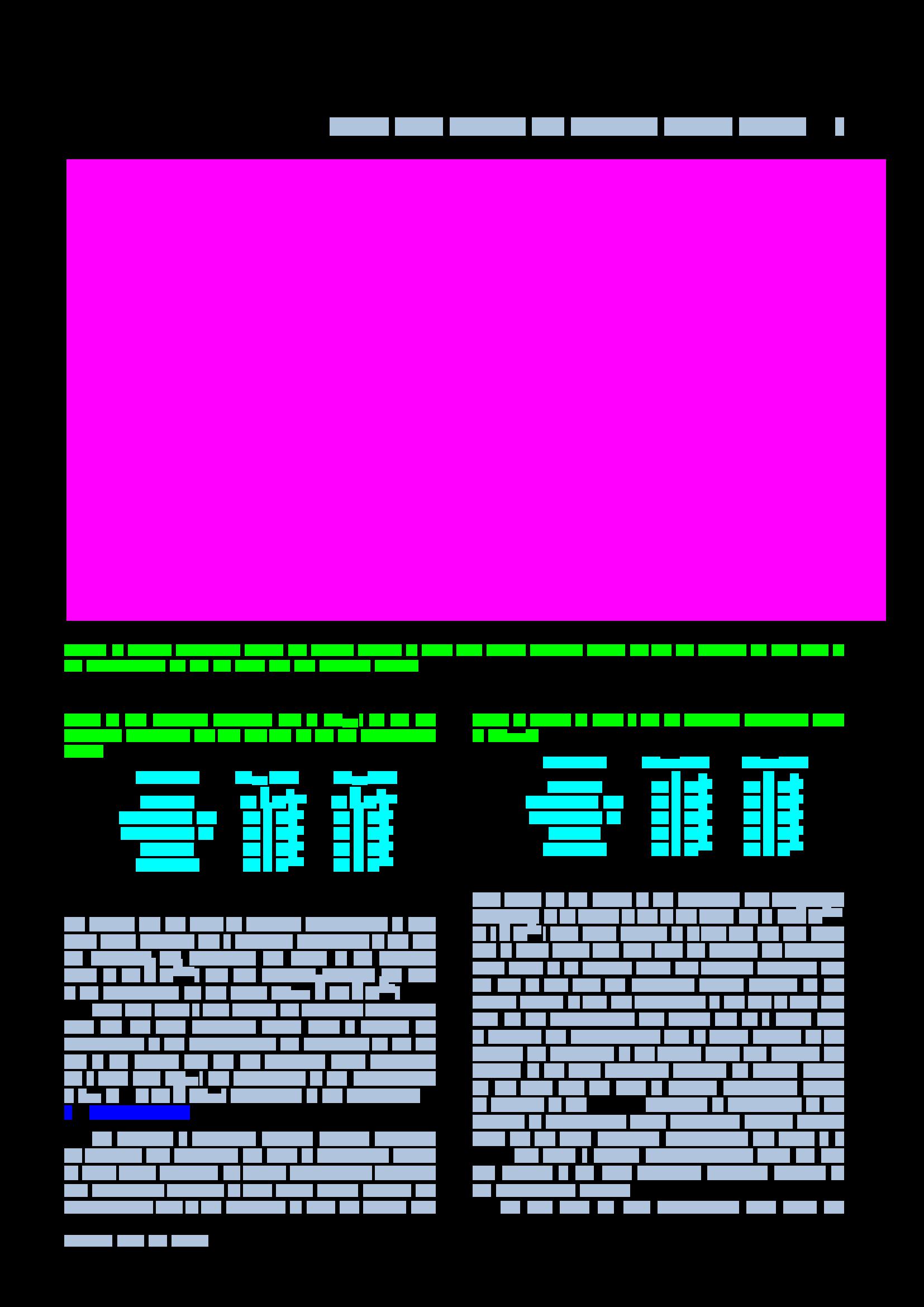}
        \caption{}
        \label{fig:3d}
    \end{subfigure}
    \caption{Example annotations of the DocBank. The colors of semantic structure labels are:  \colorbox{abstractcolor}{Abstract}, \colorbox{authorcolor}{\textcolor{white}{Author}},
    \colorbox{captioncolor}{Caption},
    \colorbox{equationcolor}{\textcolor{white}{Equation}},
    \colorbox{figurecolor}{\textcolor{white}{Figure}}, 
    \colorbox{footercolor}{Footer}, 
    \colorbox{listcolor}{List},
    \colorbox{paracolor}{Paragraph}, \colorbox{refcolor}{Reference},
    \colorbox{sectioncolor}{\textcolor{white}{Section}},   \colorbox{tablecolor}{Table}, 
    \colorbox{titlecolor}{Title}
    }\label{fig:gt-example}
\end{figure*}

To this end, we build the DocBank dataset, a document-level benchmark that contains 500K document pages with fine-grained token-level annotations for layout  analysis. Distinct from the conventional human-labeled datasets, our approach obtains high quality annotations in a simple yet effective way with weak supervision. Inspired by existing document layout annotations~\cite{Siegel2018ExtractingSF,li2019tablebank,Zhong2019PubLayNetLD}, there are a great number of digital-born documents such as the PDFs of research papers that are compiled by \LaTeX{} using their source code. The \LaTeX{} system contains the explicit semantic structure information using mark-up tags as the building blocks, such as abstract, author, caption, equation, figure, footer, list, paragraph, reference, section, table and title. To distinguish individual semantic structures, we manipulate the source code to specify different colors to the text of different semantic units. In this way, different text zones can be clearly segmented and identified as separate logical roles, which is shown in Figure~\ref{fig:gt-example}.
The advantage of DocBank is that, it can be used in any sequence labeling models from the NLP perspective. Meanwhile, DocBank can also be easily converted into image-based annotations to support object detection models in computer vision. In this way, models from different modalities can be compared fairly using DocBank, and multi-modal approaches will be further investigated and boost the performance of document layout analysis. To verify the effectiveness of DocBank, we conduct experiments using 
four baseline models: 1) BERT~\cite{bert}, a pre-trained model using only textual information based on the Transformer architecture. 2) RoBERTa~\cite{liu2019roberta}, a robustly optimized method for pre-training the Transformer architecture. 3) LayoutLM~\cite{Xu2019LayoutLMPO}, a multi-modal architecture that integrates both the text information and layout information. 4) Faster R-CNN~\cite{FasterRCNN}, a high performance object detection networks depending on region proposal algorithms to hypothesize object locations.
The experiment results show that the LayoutLM model significantly outperforms the BERT and RoBERTa models and the object detection model on DocBank for document layout analysis. We hope DocBank will empower more document layout analysis models, meanwhile promoting more customized network structures to make substantial advances in this area.

The contributions of this paper are summarized as follows:
\begin{itemize}
    \item We present DocBank, a large-scale dataset that is constructed using a weak supervision approach. It enables models to integrate both the textual and layout information for downstream tasks.
    \item We conduct a set of experiments with different baseline models and parameter settings, which confirms the effectiveness of DocBank for document layout analysis.
    \item The DocBank dataset is available at \url{https://github.com/doc-analysis/DocBank}.
\end{itemize}

\section{Task Definition}
The document layout analysis task is to extract the pre-defined semantic units in  visually rich documents. Specifically, given a document $\mathcal{D}$ composed of discrete token set $t=\{t_0,t_1, ..., t_n\}$,  each token $t_i=(w, (x_0, y_0, x_1, y_1))$ consists of word $w$ and its bounding box $(x_0, y_0, x_1, y_1)$. And $\mathcal{C} = \{c_0, c_1, .., c_m\}$ defines the semantic categories that the tokens are classified into. We intend to find a function $F:(\mathcal{C},\mathcal{D})\rightarrow \mathcal{S}$, where $\mathcal{S}$ is the prediction set:
\begin{equation}
\mathcal{S} = \{
(\{t_0^0, ..., t_0^{n_0} \}, c_0),
..., 
(\{t_k^0, ..., t_k^{n_k}  \}, c_k)
\}
\end{equation}

\section{DocBank}

\begin{figure*}[t]
\centering
\includegraphics[width=0.9\textwidth]{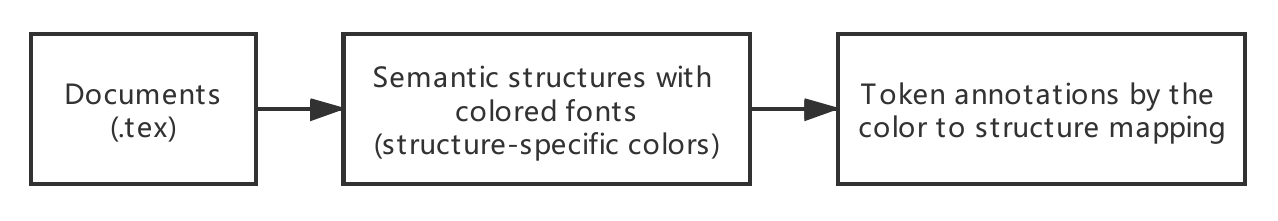}
\caption{Data processing pipeline}
\label{data-processing}
\end{figure*}

We build DocBank with token-level annotations that supports both NLP and computer vision models. As shown in Figure~\ref{data-processing}, the construction of DocBank has three steps: Document Acquisition, Semantic Structures Detection, Token Annotation. Meanwhile, DocBank can be converted to the format that is used by computer vision models in a few steps. The current DocBank dataset totally includes 500K document pages, where the training set includes 400K document pages and both the validation set and the test set include 50K document pages. 


\subsection{Document Acquisition}

We download the PDF files on arXiv.com as well as the \LaTeX{} source files since we need to modify the source code to detect the semantic structures.
The papers contain Physics, Mathematics, Computer Science and many other areas, which is beneficial for the diversity of DocBank to produce robust models. We focus on English documents in this work and will expand to other languages in the future.




\subsection{Semantic Structures Detection}\label{detection}
DocBank is a natural extension of the TableBank dataset~\cite{li2019tablebank}, where other semantic units are also included for document layout analysis. In this work, the following semantic structures are annotated in DocBank: \{Abstract, Author, Caption, Equation, Figure, Footer, List, Paragraph, Reference, Section, Table and Title\}. 
In TableBank, the tables are labeled with the help of the `fcolorbox' command. However, for DocBank, the target structures are mainly composed of text, where the `fcolorbox' cannot be well applied. Therefore, we use the `color' command to distinguish these semantic structures by changing their font colors into structure-specific colors. Basically, there are two types of commands to represent semantic structures. Some of the \LaTeX{} commands are simple words preceded by a backslash. For instance, the section titles in \LaTeX{} documents are usually in the format as follows:
\begin{center}
\begin{tabular}{|c|}
\hline
\begin{lstlisting}
\section{The title of this section}
\end{lstlisting}
\\\hline
\end{tabular}
\end{center}
Other commands often start an environment. For instance, the list declaration in \LaTeX{} documents is shown as follows:
\begin{center}
\begin{tabular}{|c|}
\hline
\begin{lstlisting}
\begin{itemize}
    \item First item
    \item Second item
\end{itemize}
\end{lstlisting}
\\\hline
\end{tabular}
\end{center}
The command $\backslash begin\{itemize\}$ starts an environment while the command $\backslash end\{itemize\}$ ends that environment. The real command name is declared as the parameters of the `begin' command and the `end' command. 

We insert the `color' command to the code of the semantic structures as follows and re-compile the \LaTeX{} documents. Meanwhile, we also define specific colors for all the semantic structures to make them distinguishable. Different structure commands require the `color' command to be placed in different locations to take effect. Finally, we get updated PDF pages from \LaTeX{} documents, where the font color of each target structure has been modified to the structure-specific color.
\begin{center}
\begin{tabular}{|c|}
\hline
\begin{lstlisting}
\section{{\color{fontcolor}{The title of this section}}}
\end{lstlisting}
\\\hline
\hline
\begin{lstlisting}
{\color{fontcolor}{\title{The title of this article}}}
\end{lstlisting}
\\\hline
\end{tabular}
\end{center}

\begin{center}
\begin{tabular}{|c||c|}
\hline
\begin{lstlisting}
\begin{itemize}
{\color{fontcolor}{
    \item First item
    \item Second item
}}
\end{itemize}
\end{lstlisting}
&
\begin{lstlisting}
{\color{fontcolor}{
    \begin{equation}
        ...
        ...
    \end{equation}
}}
\end{lstlisting}
\\\hline
\end{tabular}
\end{center}




\subsection{Token Annotation}
We use PDFPlumber\footnote{\url{https://github.com/jsvine/pdfplumber}}, a PDF parser built on PDFMiner\footnote{\url{https://github.com/euske/pdfminer}}, to extract text lines and non-text elements with their bounding boxes. Text lines are tokenized simply by white spaces, and the bounding boxes are defined as the most upper-left coordinate of characters and the most lower-right coordinate of characters, since we can only get the coordinates of characters instead of the whole tokens from the parser.
For the elements without any texts such as figures and lines in PDF files, we use the class name inside PDFMiner and wrap it using two ``\#'' symbols into a special token. The class names include ``LTFigure'' and ``LTLine'' that represent figures and lines respectively. 

The RGB values of characters and the non-text elements can be extracted by PDFPlumber from the PDF files. Mostly, a token is composed of characters with the same color. Otherwise, we use the color of the first characters as the color of the token. We determine the labels of the tokens according to the color-to-structure mapping in the Section~\ref{detection}.
A structure may contain both text and not-text elements. For instance, tables consist of words and lines. In this work, both words and lines will be annotated as the ``table'' class, so as to obtain the layout of a table as much as possible after the elements are tokenized.

\subsection{Object Detection Annotation}
\label{od_annotation}
The DocBank can be easily converted to the annotation format of the object detection models, like Faster R-CNN. The object detection models accept document images, the bounding boxes of semantic structures as input. 

We classify all the token by the type of semantic structures on a page of the document. For the tokens of the same label, we use breadth-first search to find the Connected Component. We set an x-threshold and a y-threshold. If both of the x coordinates and the y coordinates of two tokens are within the thresholds, they are ``Connected''. The breadth-first search is used to find all the tokens are connected to each other, which form an object of this label. Repeat the above steps to find all the objects. The bounding box of an object is determined by the most boundary tokens.

\subsection{Dataset Statistics}
The DocBank dataset consists of 500K document pages with 12 types of semantic units. Table~\ref{tab:unit} provides the statistics of training, validation, and test set in DocBank, showing that the number of every semantic unit and the percentage of pages with it. As these document pages are randomly drawn to generate training, validation, and test set, the distribution of semantic units in different splits are almost consistent. 

We also show the distribution of document pages across years in Table~\ref{tab:year}. We can see that the number of papers is increasing year by year. To preserve this natural distribution, we randomly sample documents of different years to build DocBank without balancing them. 

Table~\ref{tab:compare} provides a comparison of the DocBank to the previous document layout analysis datasets, including Article Regions~\cite{soto-yoo-2019-visual}, GROTOAP2~\cite{Tkaczyk2014GROTOAP2T}, PubLayNet~\cite{Zhong2019PubLayNetLD}, and TableBank~\cite{li2019tablebank}. As shown in the table, DocBank surpasses the existing datasets in both size and number of types of semantic structures. All the listed datasets are image-based while only DocBank supports both text-based and image-based models. Meanwhile, DocBank are built automatically based on the public papers, so it is extendable, which is very rare in existing datasets.

\begin{table*}[ht]
    \centering
    \resizebox{\textwidth}{!}{
    \begin{tabular}{cccccccccccccc}
    \toprule
    Split & Abstract & Author & Caption & Equation & Figure & Footer & List & Paragraph & Reference & Section & Table & Title \\
    \midrule
    \multirow{2}{*}{Train} & 25,387 & 25,909 & 106,723 & 161,140 & 90,429 & 38,482 & 44,927 & 398,086 & 44,813 & 180,774 & 19,638 & 21,688\\
     & 6.35\% & 6.48\% & 26.68\% & 40.29\% & 22.61\% & 9.62\% & 11.23\% & 99.52\% & 11.20\% & 45.19\% & 4.91\% & 5.42\%\\
     \midrule
    \multirow{2}{*}{Validation} & 3,164 & 3,286 & 13,443 & 20,154 & 11,463 & 4,804 & 56,09 & 49,759 & 55,49 & 22,666 & 2,374 & 2,708\\
     & 6.33\% & 6.57\% & 26.89\% & 40.31\% & 22.93\% & 9.61\% & 11.22\% & 99.52\% & 11.10\% & 45.33\% & 4.75\% & 5.42\%\\
     \midrule
    \multirow{2}{*}{Test} & 3,176 & 3,277 & 13,476 & 20,244 & 11,378 & 4,876 & 5,553 & 49,762 & 5,641 & 22,384 & 2,505 & 2,729\\
     & 6.35\% & 6.55\% & 26.95\% & 40.49\% & 22.76\% & 9.75\% & 11.11\% & 99.52\% & 11.28\% & 44.77\% & 5.01\% & 5.46\%\\
     \midrule
    \multirow{2}{*}{All} & 31,727 & 32,472 & 133,642 & 201,538 & 113,270 & 48,162 & 56,089 & 497,607 & 56,003 & 225,824 & 24,517 & 27,125\\
     & 6.35\% & 6.49\% & 26.73\% & 40.31\% & 22.65\% & 9.63\% & 11.22\% & 99.52\% & 11.20\% & 45.16\% & 4.90\% & 5.43\%\\
     \bottomrule
    \end{tabular}
    }
    \caption{Semantic Structure Statistics of training, validation, and test sets in DocBank}
    \label{tab:unit}
\end{table*}

\begin{table*}[ht]
\centering
\small
\begin{tabular}{ccccccccc}
\toprule
Year & \multicolumn{2}{c}{Train} & \multicolumn{2}{c}{Validation} & \multicolumn{2}{c}{Test} & \multicolumn{2}{c}{All} \\
\midrule
2014 & 65,976 & 16.49\% & 8,270 & 16.54\% & 8,112 & 16.22\% & 82,358 & 16.47\% \\
2015 & 77,879 & 19.47\% & 9,617 & 19.23\% & 9,700 & 19.40\% & 97,196 & 19.44\% \\
2016 & 87,006 & 21.75\% & 10,970 & 21.94\% & 10,990 & 21.98\% & 108,966 & 21.79\% \\
2017 & 91,583 & 22.90\% & 11,623 & 23.25\% & 11,464 & 22.93\% & 114,670 & 22.93\% \\
2018 & 77,556 & 19.39\% & 9,520 & 19.04\% & 9,734 & 19.47\% & 96,810 & 19.36\% \\
Total & 400,000 & 100.00\% & 50,000 & 100.00\% & 50,000 & 100.00\% & 500,000 & 100.00\% \\
\bottomrule
\end{tabular}
\caption{Year Statistics of training, validation, and test sets in DocBank}
\label{tab:year}
\end{table*}

\begin{table*}[ht]
    \centering
    \small
    \begin{tabular}{ccccccc}
    \toprule
     \bf Dataset & \bf \#Pages & \bf \#Units & \bf Image-based? & \bf Text-based? & \bf Fine-grained? & \bf Extendable?\\
     \midrule
     Article Regions & 100 & 9 & \ding{51} & \ding{55} & \ding{51} & \ding{55}\\
     GROTOAP2   & 119,334 & 22 & \ding{51} & \ding{55} & \ding{55} & \ding{55}\\
     PubLayNet   & 364,232 & 5 & \ding{51} & \ding{55} & \ding{51} & \ding{55}\\
     TableBank   & 417,234 & 1 & \ding{51} & \ding{55} & \ding{51} & \ding{51}\\
     DocBank     & 500,000 & 12 & \ding{51} & \ding{51} & \ding{51} & \ding{51}\\
    \bottomrule
    \end{tabular}
    \caption{Comparison of DocBank with existing document layout analysis datasets}
    \label{tab:compare}
\end{table*}

\section{Method}


As the dataset was fully annotated at token-level, we consider the document layout analysis task as a text-based sequence labeling task. 
Under this setting, we evaluate three representative pre-trained language models on our dataset including BERT, RoBERTa and LayoutLM to validate the effectiveness of DocBank. To verify the performance of the models from different modalities on DocBank, we train the Faster R-CNN model on the object detection format of DocBank and unify its output with the sequence labeling models to evaluate.

\subsection{Models}
\paragraph{The BERT Model}
BERT is a Transformer-based language model trained on large-scale text corpus. It consists of a multi-layer bidirectional Transformer encoder. It accepts a token sequence as input and calculates the input representation by summing the corresponding token, segment, and position embeddings. Then, the input vectors pass multi-layer attention-based Transformer blocks to get the final contextualized language representation.



\paragraph{The RoBERTa Model}
RoBERTa~\cite{liu2019roberta} is a more powerful version of BERT, which has been proven successfully in many NLP tasks. Basically, the model architecture is the same as BERT except for the tokenization algorithm and improved training strategies. By increasing the size of the pre-training data and the number of training steps, RoBERTa gets better performance on several downstream tasks.


\paragraph{The LayoutLM Model}
LayoutLM is a multi-modal pre-trained language model that jointly models the text and layout information of visually rich documents. In particular, it has an additional 2-D position embedding layer to embed the spatial position coordinates of elements. In detail, the LayoutLM model accepts a sequence of tokens with corresponding bounding boxes in documents. Besides the original embeddings in BERT, LayoutLM feeds the bounding boxes into the additional 2-D position embedding layer to get the layout embeddings. Then the summed representation vectors pass the BERT-like multi-layer Transformer encoder. Note that we use the LayoutLM without image  embeddings and more details are provided in the Section~\ref{pretraining}.



\paragraph{The Faster R-CNN Model}
Faster R-CNN is one of the most popular object detection networks. It proposes the Region Proposal Network (RPN) to address the bottleneck of region proposal computation. RPN shares convolutional features with the detection network using `attention' mechanisms, which leads to nearly cost-free region proposals and high accuracy on many object detection benchmarks.

\subsection{Pre-training LayoutLM}
\label{pretraining}
LayoutLM chooses the Masked Visual-Language Model(MVLM) and Multi-label Document Classiﬁcation(MDC) as the objectives when pre-training the model. For the MVLM task,its procedure is to simply mask some of the input tokens at random keeping the corresponding position embedding and then predict those masked tokens. In this case, the final hidden vectors corresponding to the mask tokens are fed into an output softmax over the vocabulary. For the MDC task, it uses the output context vector of [CLS] token to predict the category labels. With these two training objectives, the LayoutLM is pre-trained on IIT-CDIP Test Collection 1.0\footnote{\url{https://ir.nist.gov/cdip/}}~\cite{Lewis:2006:BTC:1148170.1148307}, a large document image collection.




\subsection{Training Samples in Reading Order}
We organize the DocBank dataset using the reading order, which means that we sort all the text boxes (a hierarchy level higher than text line in PDFMiner) and non-text elements from top to bottom by their top border positions. The text lines inside a text box are already sorted top-to-bottom. We tokenize all the text lines in the left-to-right order and annotate them. Basically, all the tokens are arranged top-to-bottom and left-to-right, which is also applied to all the columns of multi-column documents.


\subsection{Fine-tuning} 

We fine-tune the pre-trained model with the DocBank dataset. As the document layout analysis is regarded as a sequence labeling task, all the tokens are labeled using the output with the maximum probability. The number of output class equals the number of semantic structure types.


\section{Experiment}


\subsection{Evaluation Metrics}
\label{metrics}


As the inputs of our model are serialized 2-D documents, the typical BIO-tagging evaluation is not suitable for our task. The tokens of each semantic unit may discontinuously distribute in the input sequence.
In this case, we proposed a new metric, especially for text-based document layout analysis methods. For each kind of document semantic structure, we calculated their metrics individually. The definition is as follows:

\begin{small}
\begin{gather*}
Precision=\frac{\text{Area of Ground truth tokens in Detected tokens}}{\text{Area of all Detected tokens}},\notag \\
Recall=\frac{\text{Area of Ground truth tokens in Detected tokens}}{\text{Area of all Ground truth tokens}},\notag \\
F1\text{ }Score=\frac{\text{2} \times \text{Precision} \times \text{Recall}}{\text{Precision + Recall}}\notag.
\end{gather*}  
\end{small}

\subsection{Settings}
Our baselines of BERT and RoBERTa are built upon the HuggingFace's Transformers~\cite{Wolf2019HuggingFacesTS} while the LayoutLM baselines are implemented with the codebase in LayoutLM's official repository\footnote{\url{https://aka.ms/layoutlm}}. We use 8 V100 GPUs with a batch size of 10 per GPU. It takes 5 hours to fine-tune 1 epoch on the 400K document pages. We use the BERT and RoBERTa tokenizers to tokenize the training samples and optimized the model with AdamW. The initial learning rate of the optimizer is $5\times10^{-5}$. We split the data into a max block size of $N=512$. We use the Detectron2~\cite{wu2019detectron2} to train the Faster R-CNN model on DocBank. We use the Faster R-CNN algorithm with the ResNeXt~\cite{resnext} as the backbone network architecture, where the parameters are pre-trained on the ImageNet dataset.

\subsection{Results}

\begin{table*}[ht]
    \centering
    \resizebox{\textwidth}{!}{

    \begin{tabular}{c|cccccccccccc|c}
    \toprule
    Models & Abstract & Author & Caption & Equation & Figure & Footer & List & Paragraph & Reference & Section & Table & Title & Macro average \\
    \midrule
    $\textrm{BERT}_{\rm BASE}$ & 0.9294 & 0.8484 & 0.8629 & 0.8152 & 1.0000 & 0.7805 & 0.7133 & 0.9619 & 0.9310 & 0.9081 & 0.8296 & 0.9442 & 0.8770 \\
    $\textrm{RoBERTa}_{\rm BASE}$ & 0.9288 & 0.8618 & 0.8944 & 0.8248 & 1.0000 & 0.8014 & 0.7353 & 0.9646 & 0.9341 & 0.9337 & 0.8389 & 0.9511 & 0.8891 \\
    $\textrm{LayoutLM}_{\rm BASE}$ & \textbf{0.9816} & 0.8595 & 0.9597 & 0.8947 & 1.0000 & 0.8957 & 0.8948 & 0.9788 & 0.9338 & 0.9598 & 0.8633 & \textbf{0.9579} & 0.9316 \\
    $\textrm{BERT}_{\rm LARGE}$ & 0.9286 & 0.8577 & 0.8650 & 0.8177 & 1.0000 & 0.7814 & 0.6960 & 0.9619 & 0.9284 & 0.9065 & 0.8320 & 0.9430 & 0.8765 \\
    $\textrm{RoBERTa}_{\rm LARGE}$ & 0.9479 & 0.8724 & 0.9081 & 0.8370 & 1.0000 & 0.8392 & 0.7451 & 0.9665 & 0.9334 & 0.9407 & 0.8494 & 0.9461 & 0.8988 \\
    $\textrm{LayoutLM}_{\rm LARGE}$ & 0.9784 & 0.8783 & 0.9556 & 0.8974 & \textbf{1.0000} & 0.9146 & 0.9004 & 0.9790 & 0.9332 & 0.9596 & 0.8679 & 0.9552 & 0.9350 \\
    \midrule
    X101 & 0.9717 & 0.8227 & 0.9435 & 0.8938 & 0.8812 & 0.9029 & 0.9051 & 0.9682 & 0.8798 & 0.9412 & 0.8353 & 0.9158 & 0.9051 \\
    X101+$\textrm{LayoutLM}_{\rm BASE}$ & 0.9815 & 0.8907 & \textbf{0.9669} & 0.9430 & 0.9990 & 0.9292 & \textbf{0.9300} & 0.9843 & \textbf{0.9437} & 0.9664 & 0.8818 & 0.9575 & 0.9478 \\
    X101+$\textrm{LayoutLM}_{\rm LARGE}$ & 0.9802 & \textbf{0.8964} & 0.9666 & \textbf{0.9440} & 0.9994 & \textbf{0.9352} & 0.9293 & \textbf{0.9844} & 0.9430 & \textbf{0.9670} & \textbf{0.8875} & 0.9531 & \textbf{0.9488} \\
    \bottomrule
    \end{tabular}
    }
    \caption{The performance of BERT, RoBERTa, LayoutLM and Faster R-CNN on the DocBank test set.}
    \label{tab:result}
\end{table*}

The evaluation results of BERT, RoBERTa and LayoutLM are shown in Table~\ref{tab:result}. 
We evaluate six models on the test set of DocBank. We notice that the LayoutLM gets the highest scores on the \{abstract, author, caption, equation, figure, footer, list, paragraph, section, table, title\} labels. The RoBERTa model gets the best performance on the ``reference'' label but the gap with the LayoutLM is very small. This indicates that the LayoutLM architecture is significantly better than the BERT and RoBERTa architecture in the document layout analysis task. 

We also evaluate the ResNeXt-101 model and two ensemble models combining ResNeXt-101 and LayoutLM. The output of the ResNeXt-101 model is the bounding boxes of semantic structures. To unify the outputs of them, we mark the tokens inside each bounding box by the label of the corresponding bounding box. After that, we calculate the metrics following the equation in Section~\ref{metrics}.

\begin{figure*}[t]
\centering
    \begin{subfigure}[b]{0.235\textwidth}
        \includegraphics[width=\textwidth]{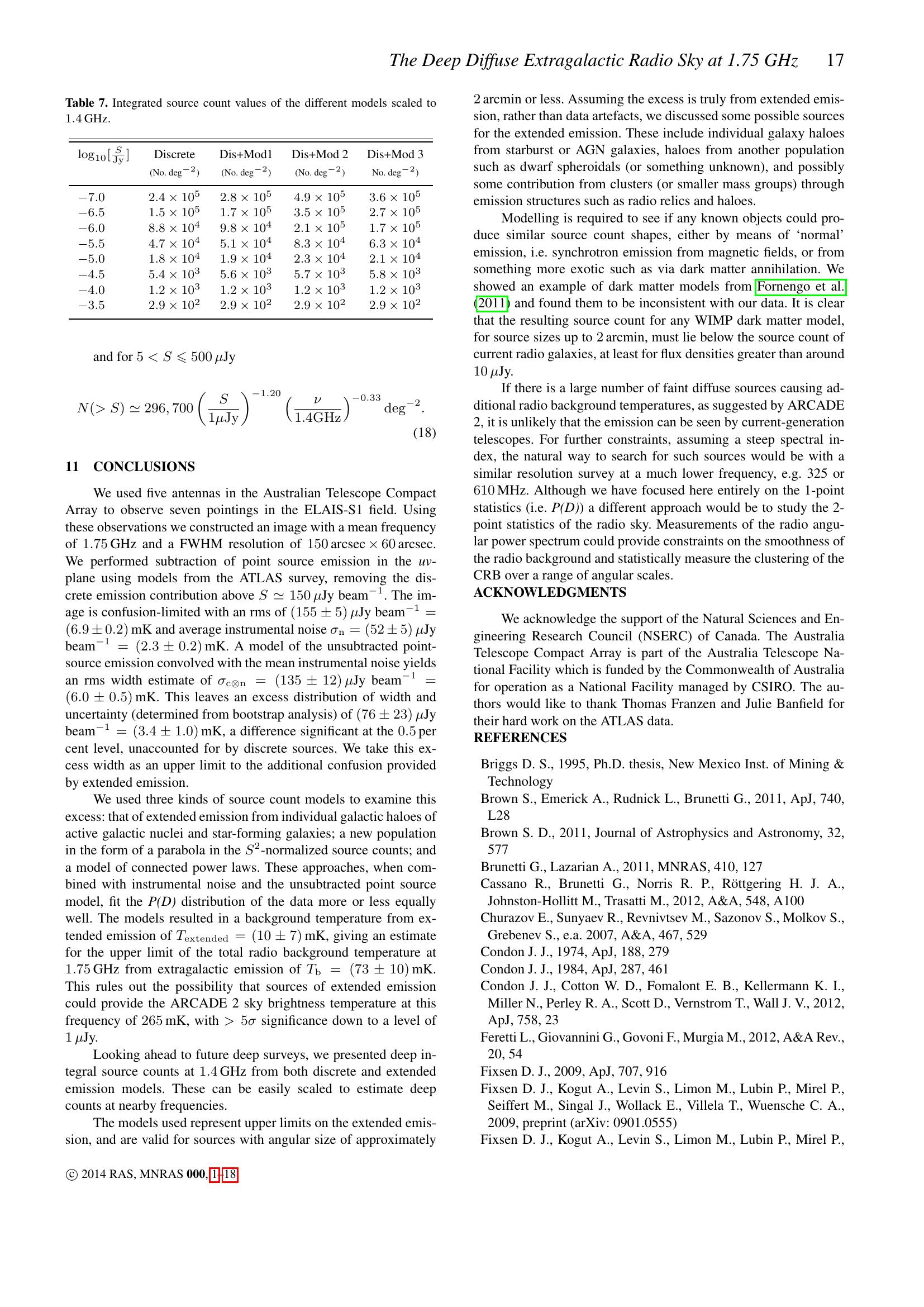}
        \caption{Original document page}
        \label{fig:nncase-a}
    \end{subfigure}
    ~ 
    \begin{subfigure}[b]{0.235\textwidth}
        \includegraphics[width=\textwidth]{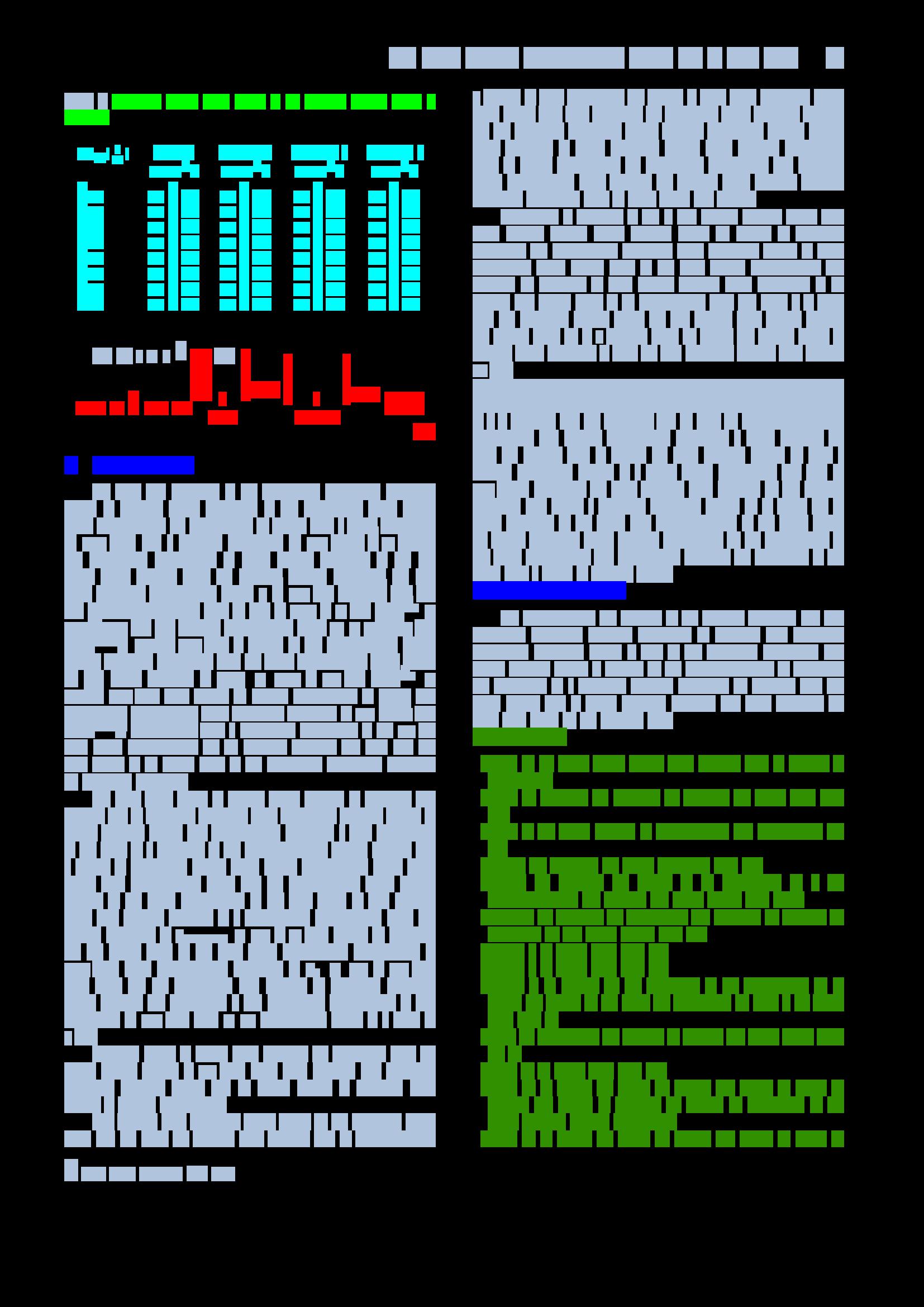}
        \caption{Groundtruth}
        \label{fig:nncase-b}
    \end{subfigure}
    ~ 
    \begin{subfigure}[b]{0.235\textwidth}
        \includegraphics[width=\textwidth]{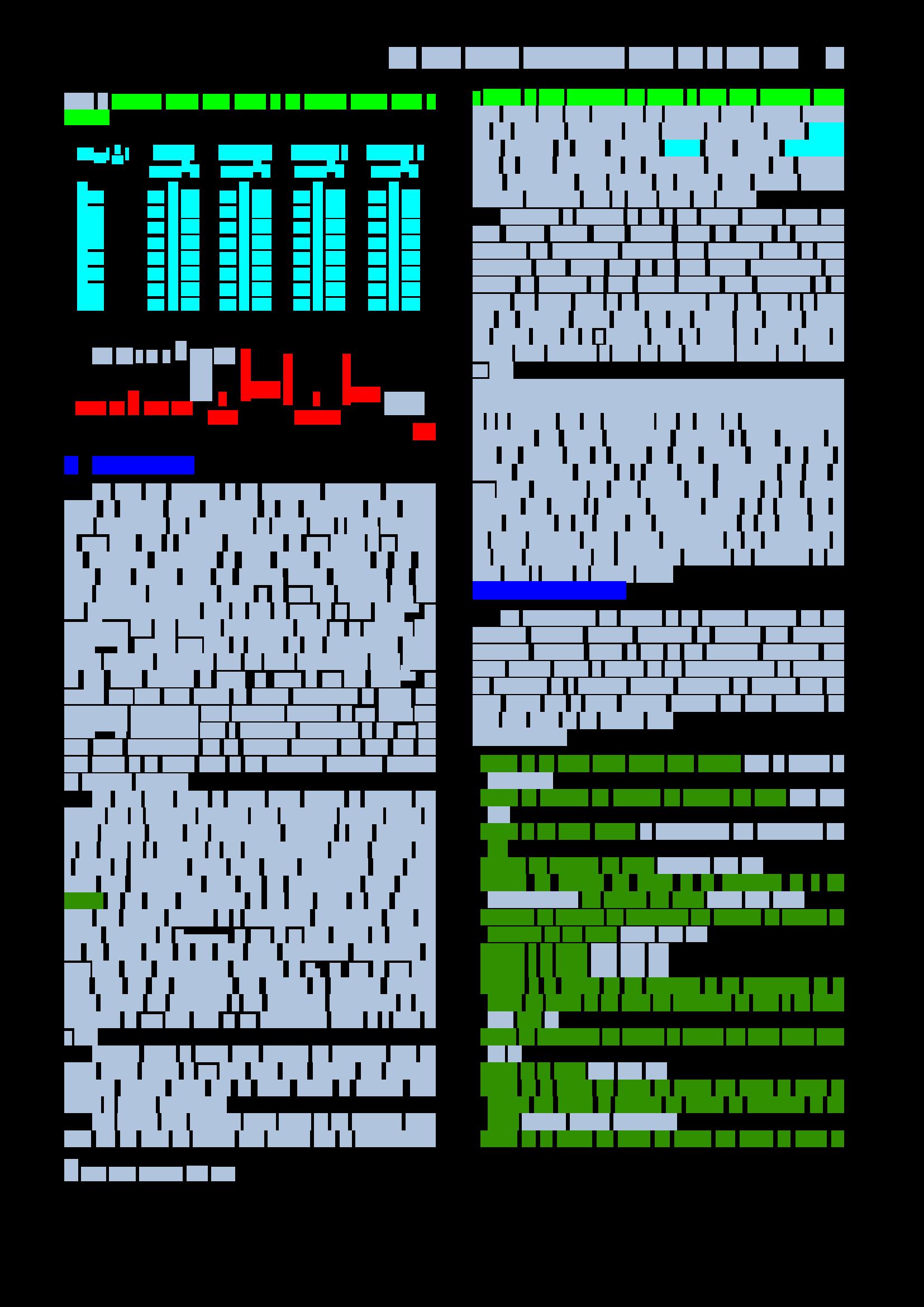}
        \caption{Pre-trained BERT}
        \label{fig:nncase-c}
    \end{subfigure}
    ~
    \begin{subfigure}[b]{0.235\textwidth}
        \includegraphics[width=\textwidth]{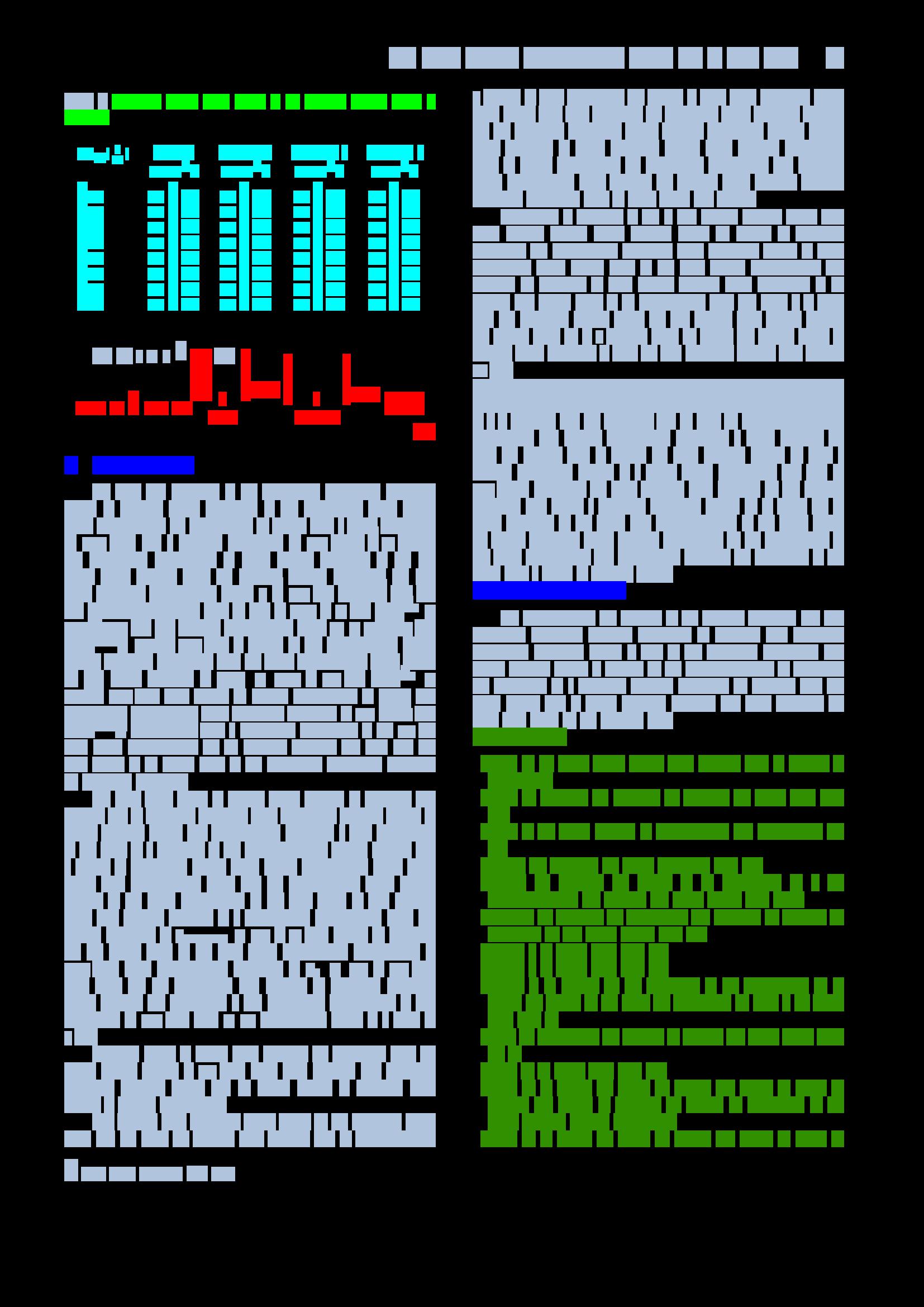}
        \caption{Pre-trained LayoutLM}
        \label{fig:nncase-d}
    \end{subfigure}
    \caption{Example output of pre-trained LayoutLM and pre-trained BERT on the test set}\label{fig:nncase}
\end{figure*}

\begin{figure*}[t]
\centering
    \begin{subfigure}[b]{0.235\textwidth}
        \includegraphics[width=\textwidth]{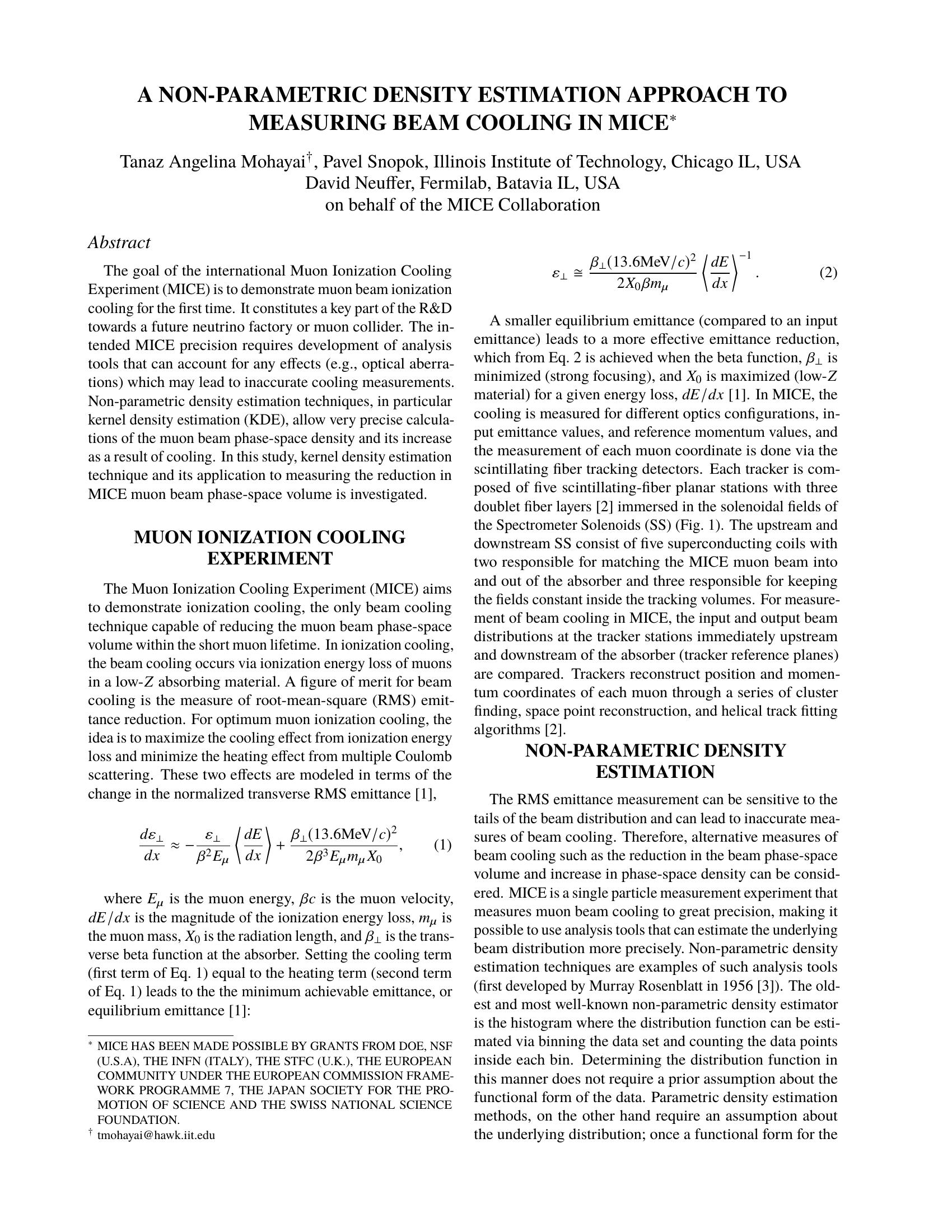}
        \caption{Original document page}
        \label{fig:rocase-a}
    \end{subfigure}
    ~ 
    \begin{subfigure}[b]{0.235\textwidth}
        \includegraphics[width=\textwidth]{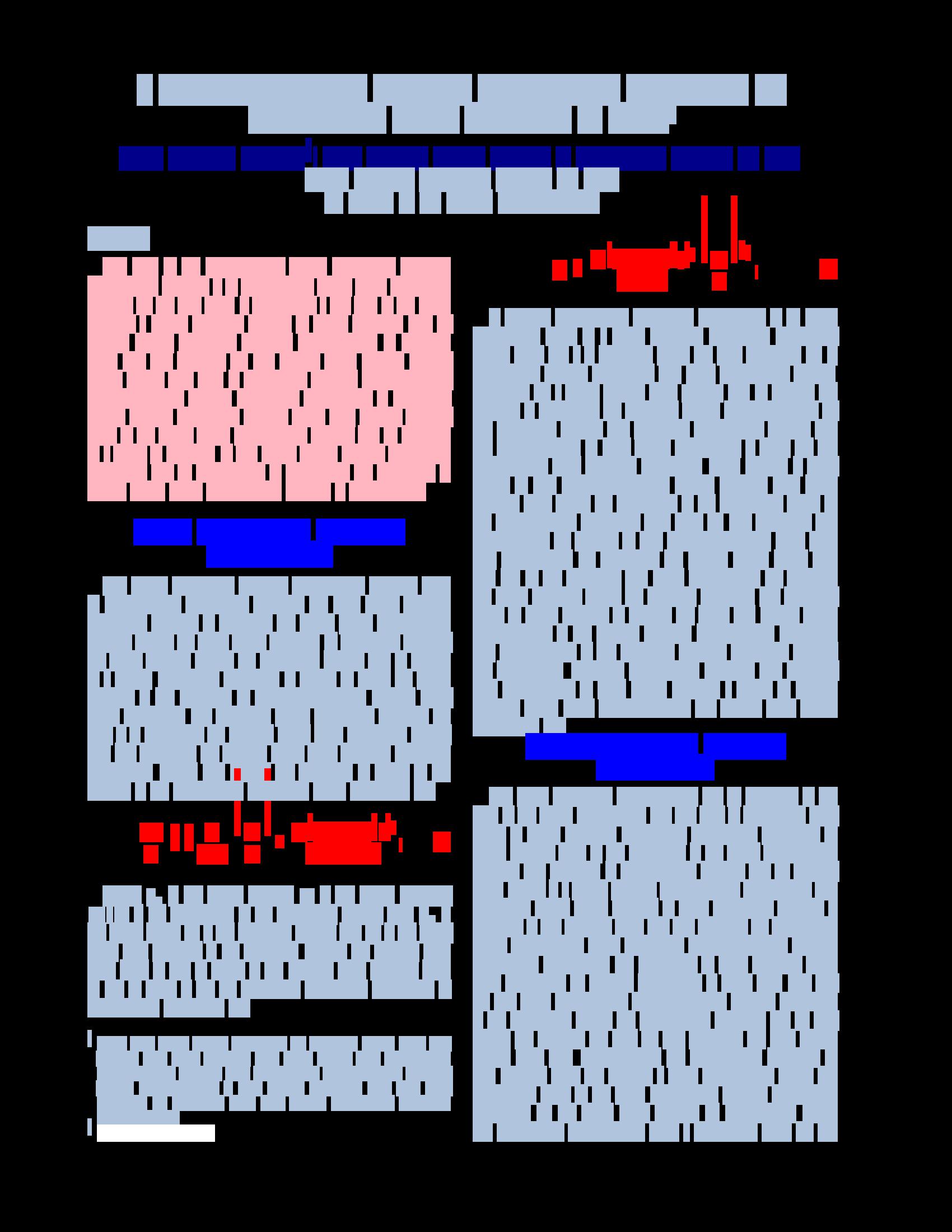}
        \caption{Groundtruth}
        \label{fig:rocase-b}
    \end{subfigure}
    ~ 
    \begin{subfigure}[b]{0.235\textwidth}
        \includegraphics[width=\textwidth]{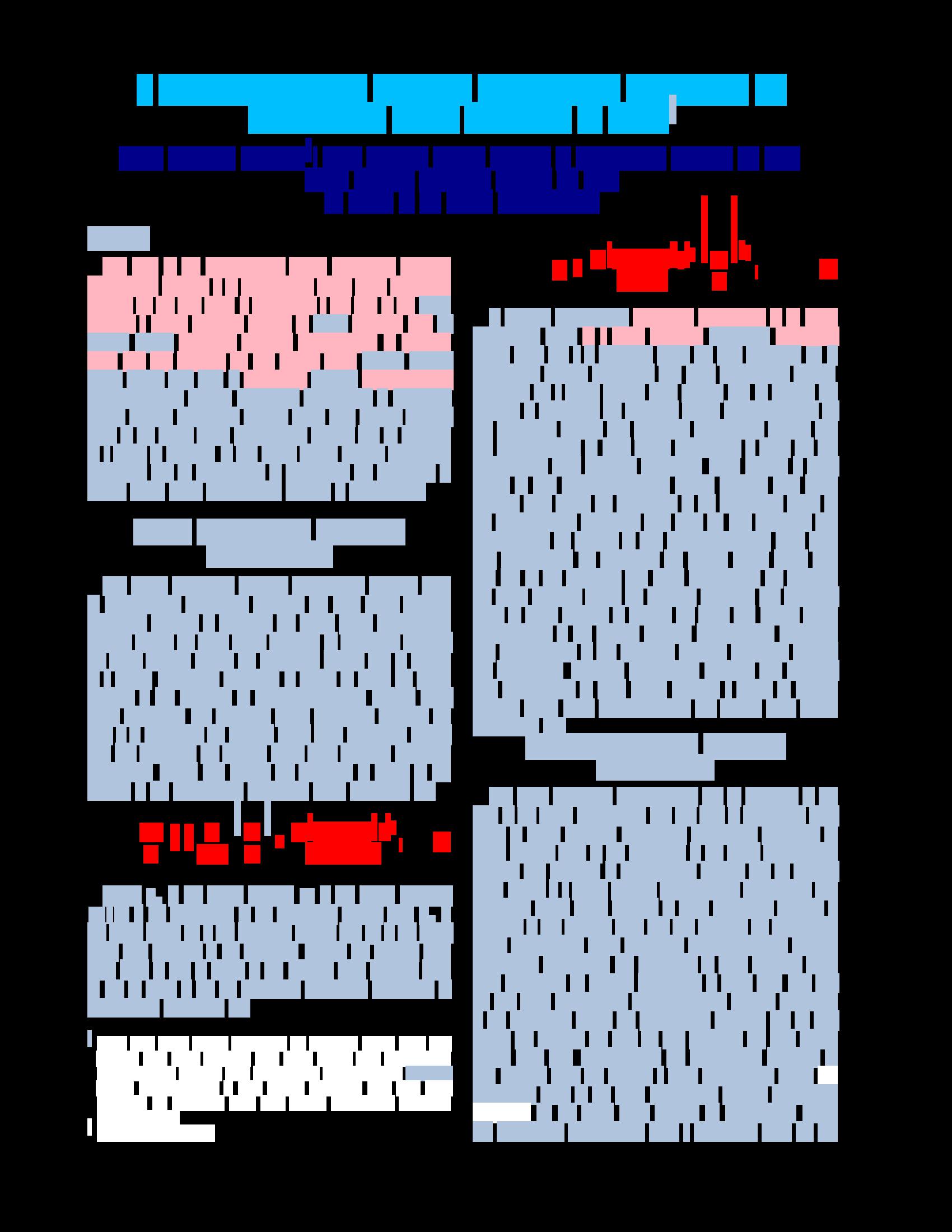}
        \caption{Pre-trained BERT}
        \label{fig:rocase-c}
    \end{subfigure}
    ~
    \begin{subfigure}[b]{0.235\textwidth}
        \includegraphics[width=\textwidth]{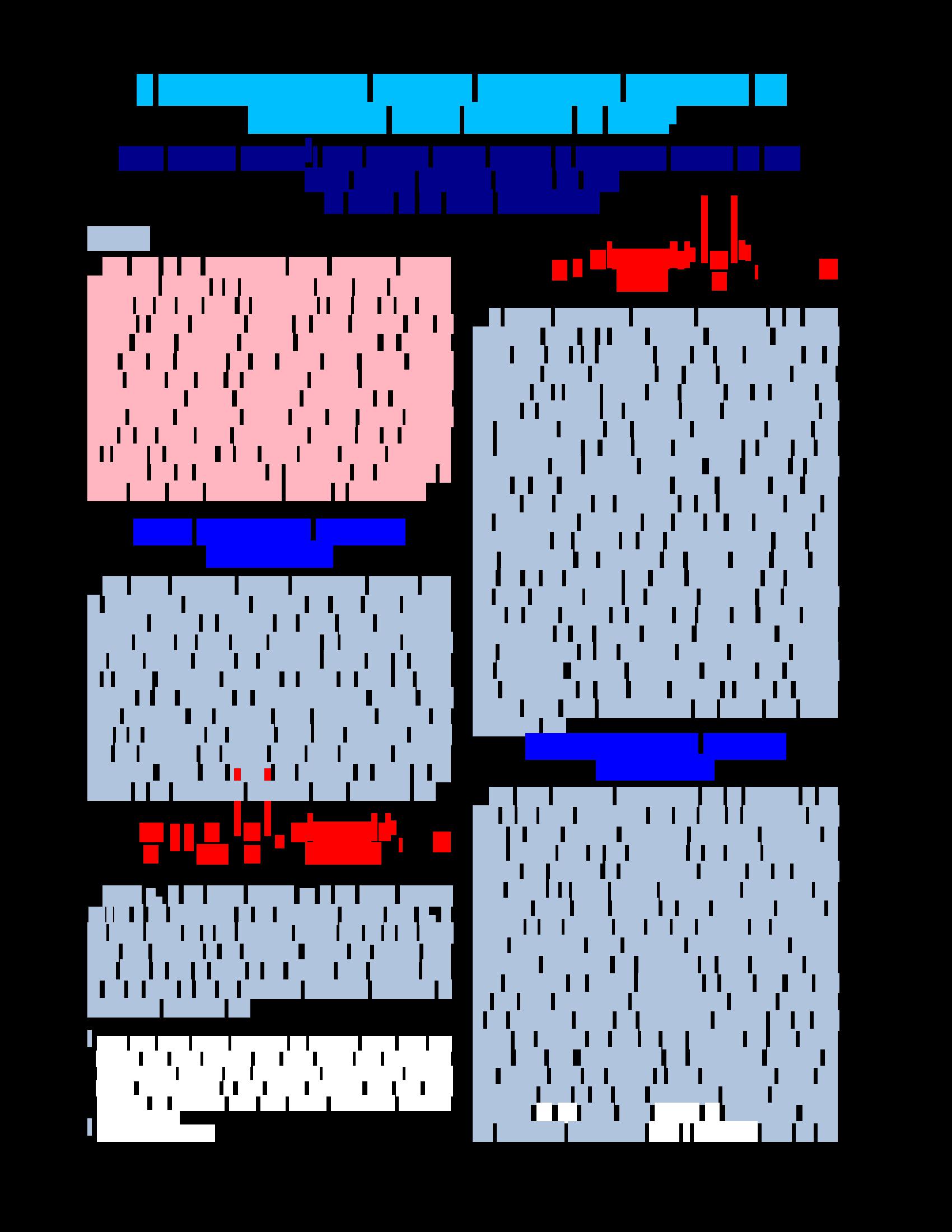}
        \caption{Pre-trained LayoutLM}
        \label{fig:rocase-d}
    \end{subfigure}
    \caption{Example output of pre-trained LayoutLM and pre-trained BERT on the test set}\label{fig:rocase}
\end{figure*}



\section{Case Study}


We visualize the outputs of pre-trained BERT and pre-trained LayoutLM on some samples of the test set in Figure~\ref{fig:nncase} and Figure~\ref{fig:rocase}. Generally, it is observed that the sequence labeling method performs well on the DocBank dataset, where different semantic units can be identified. For the pre-trained BERT model, we can see some tokens are detected incorrectly, which illustrates that only using text information is still not sufficient for document layout analysis tasks, and visual information should be considered as well. Compared with the pre-trained BERT model, the pre-trained LayoutLM model integrates both the text and layout information. Therefore, it produces much better performance on the benchmark dataset. This is because the 2D position embeddings can model spatial distance and boundary of semantic structures in a unified framework, which leads to the better detection accuracy.

\section{Related Work}

The research of document layout analysis can be divided into three categories: rule-based approaches, conventional machine learning approaches, and deep learning approaches. 

\subsection{Rule-based Approaches}

Most of the rule-based works~\cite{lebourgeois1992fast,ha1995document,simon1997fast,ha1995recursive} are divided into two main categories: the bottom-up approaches and the top-down approaches.

Some bottom-up approaches~\cite{lebourgeois1992fast,ha1995document,simon1997fast} first detect the connected components of black pixels as the basic computational units in document image analysis. The main part of the document segment process is combining them into higher-level structures through different heuristics methods and labeling them according to different structural features. 
The spatial auto-correlation approach \cite{ref76,ref77} is a bottom-up texture-based method for document layout analysis. It starts by extracting texture features directly from the image pixels to form homogeneous regions and will auto-correlate the document image with itself to highlight periodicity and texture orientation.



For the top-down strategy, \cite{ref75} proposed a mask-based texture analysis to locate text regions written in different languages. 
Run Length Smearing Algorithm converts image-background to image-foreground if the number of background pixels between any two consecutive foreground pixels is less than a predefined threshold, which is first introduced by \cite{ref162}. 
Document projection profile method was proposed to detect document regions\cite{ref142}. 
\cite{ref105} proposed a X-Y cut algorithm that used projection profile to determine document blocks cuts. 
For the above work, the rule-based heuristic algorithm is difficult to process complex documents, and the applicable document types are relatively simple.

\begin{figure*}[t]
\centering
    \begin{subfigure}[b]{0.235\textwidth}
        \includegraphics[width=\textwidth]{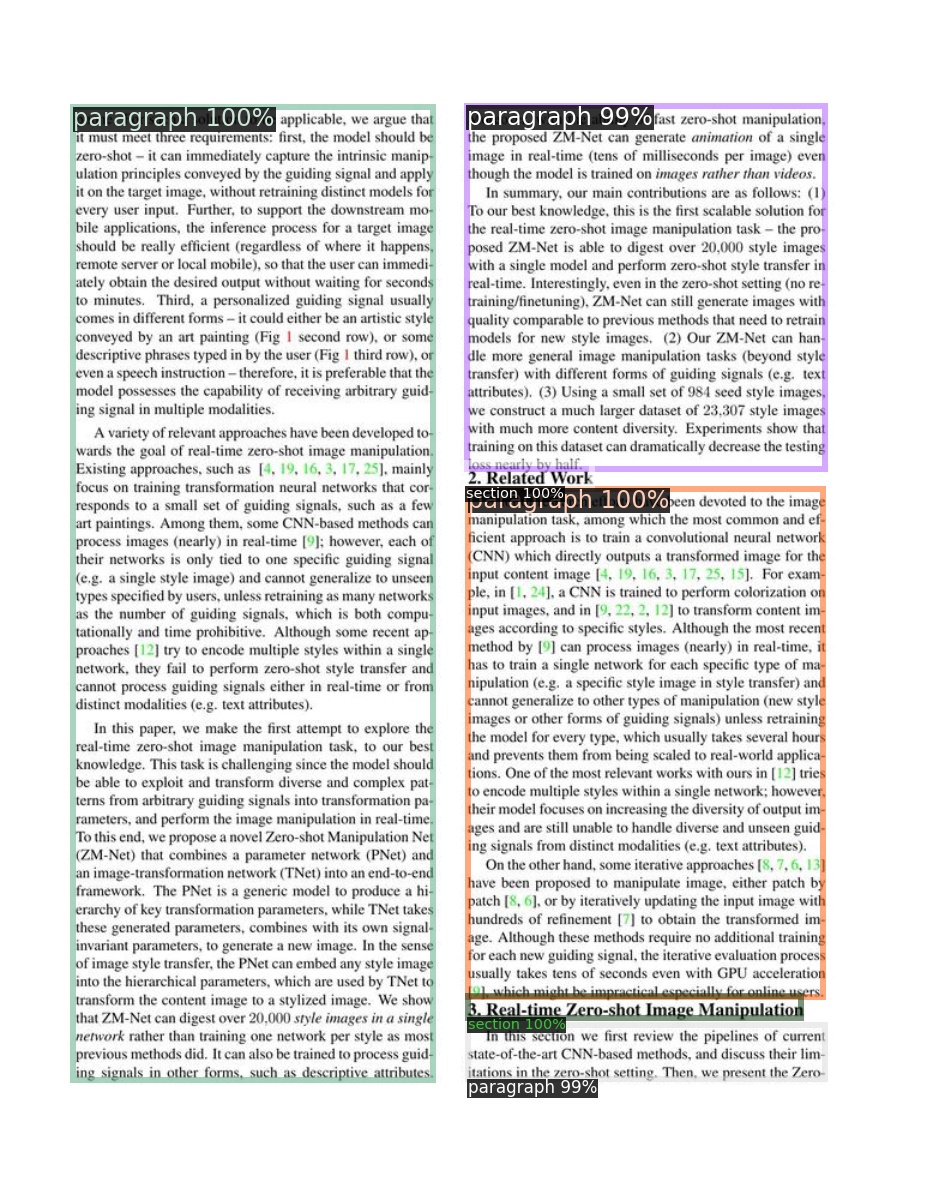}
        \label{fig:cvcase-a}
    \end{subfigure}
    ~ 
    \begin{subfigure}[b]{0.235\textwidth}
        \includegraphics[width=\textwidth]{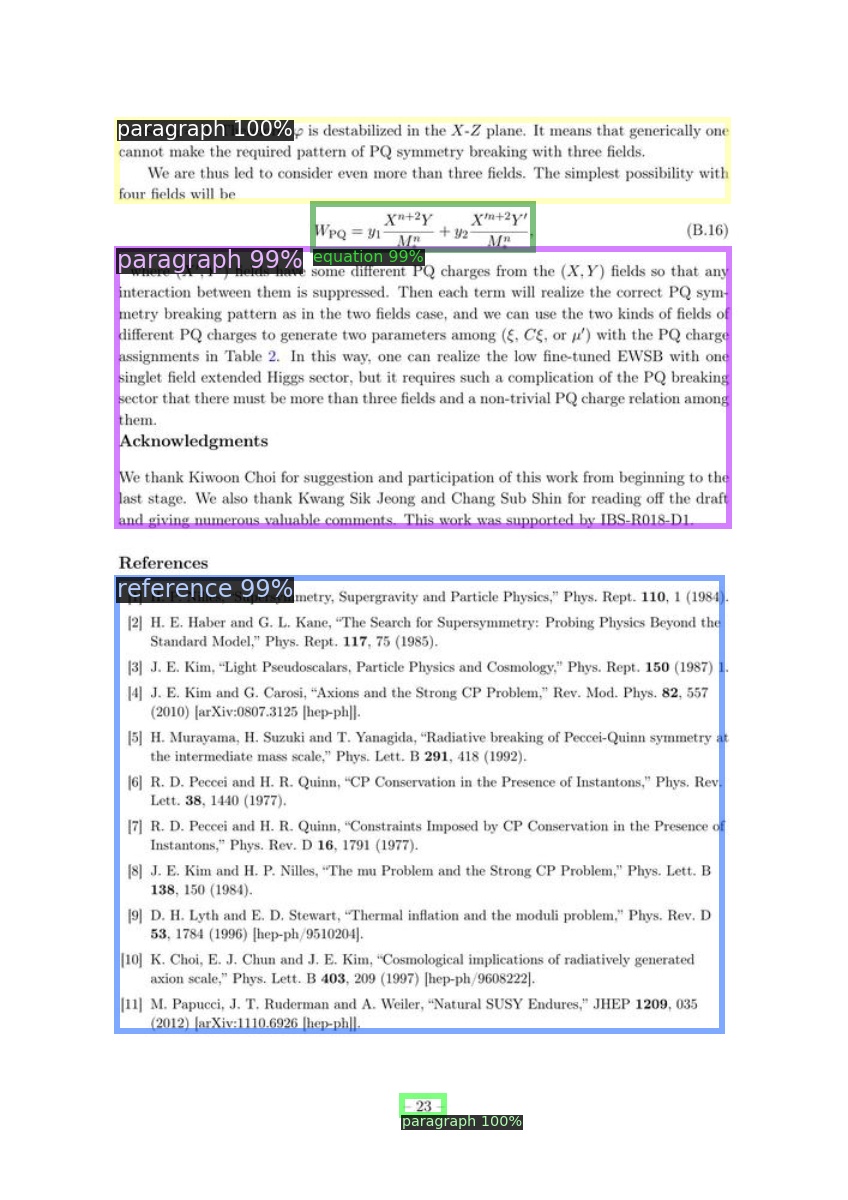}
        \label{fig:cvcase-b}
    \end{subfigure}
    ~ 
    \begin{subfigure}[b]{0.235\textwidth}
        \includegraphics[width=\textwidth]{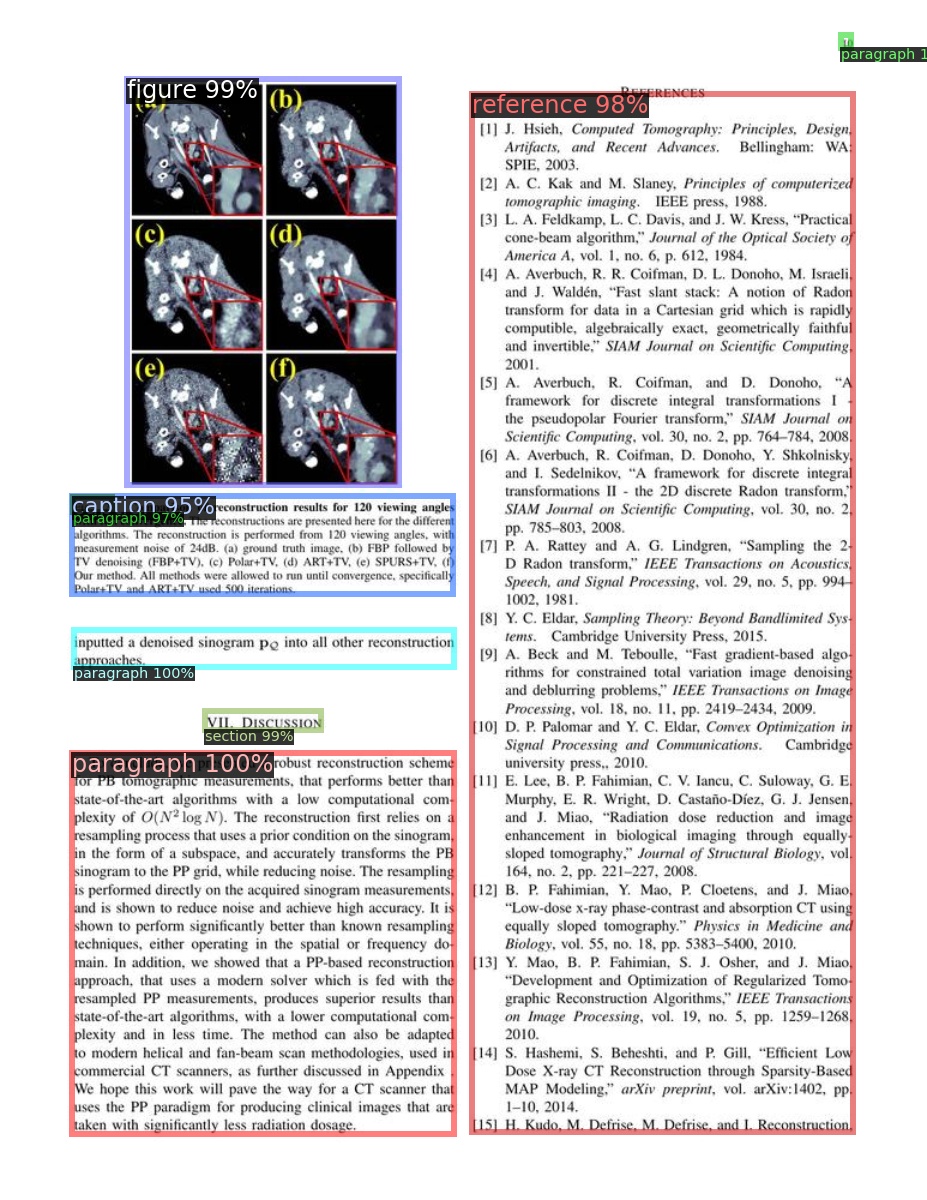}
        \label{fig:cvcase-c}
    \end{subfigure}
    ~
    \begin{subfigure}[b]{0.235\textwidth}
        \includegraphics[width=\textwidth]{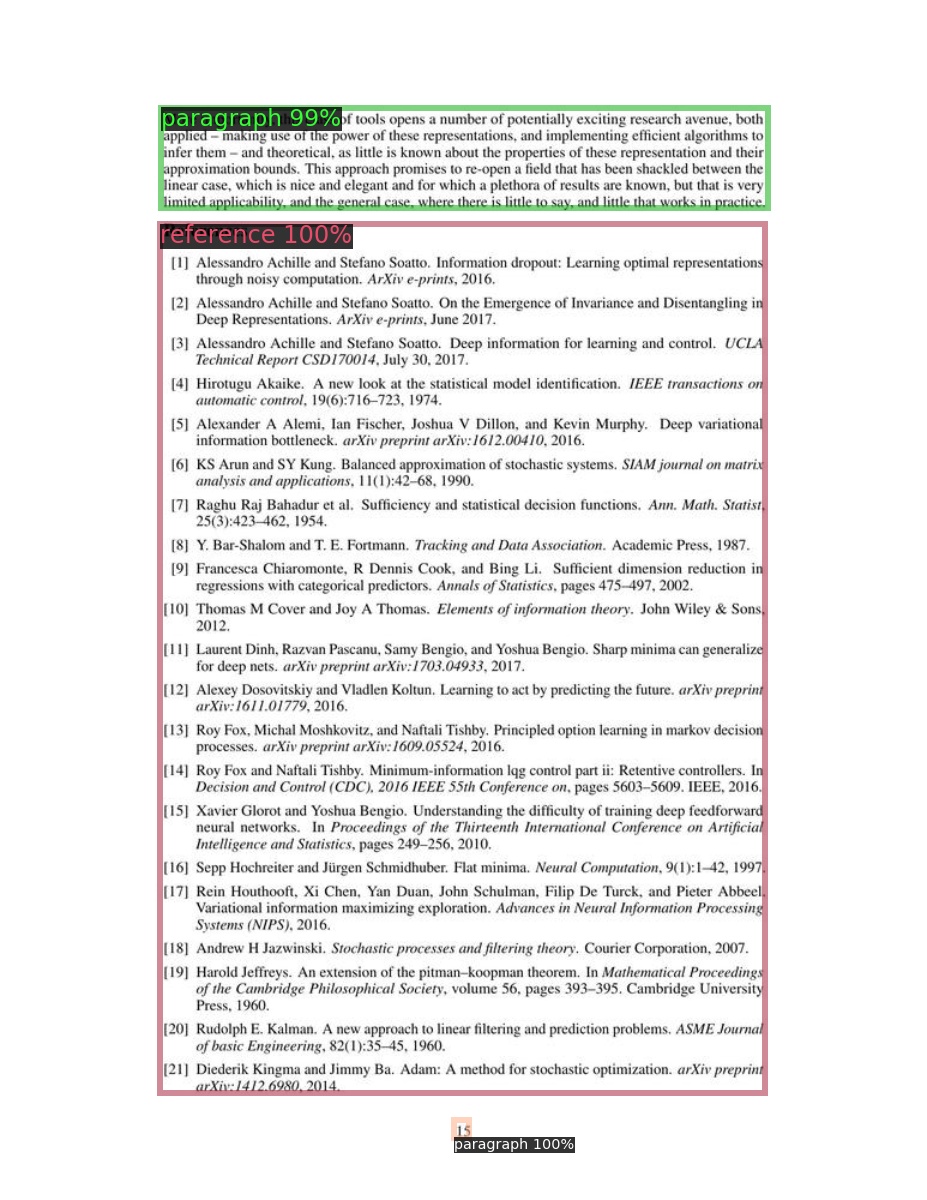}
        \label{fig:cvcase-d}
    \end{subfigure}
    \caption{Example output of Faster R-CNN on the test set}\label{fig:cvcase}
\end{figure*}

\subsection{Conventional Machine Learning Approaches}



To address the issue about data imbalance that the learning-based methods suffer from, a dynamic MLP~(DMLP) was proposed to learn a less-biased machine model using pixel-values and context information \cite{ref22}. 
Usually, block and page-based analysis require feature extraction methods to empower the training and build robust models. The handcrafted features are developed through feature extraction techniques such as Gradient Shape Feature~(GSF) \cite{ref52} or Scale Invariant Feature Transform~(SIFT) \cite{ref62,ref63,ref65,ref164}. 
There are several other techniques that use features extraction methods such as texture features \cite{ref44,ref101,ref102,ref103,ref163,ref165} and geometric features \cite{ref30,ref31}. 
Manually designing features require a large amount of work and is difficult to obtain a highly abstract semantic context. Moreover, the above machine learning methods rely solely on visual cues and ignore textual information.

\subsection{Deep Learning Approaches}

The learning-based document layout analysis methods get more attention to address complex layout analysis. \cite{ref38} suggested a Fully Convolutional Neural Network (FCNN) with a weight-training loss scheme, which was designed mainly for text-line extraction, while the weighting loss in FCNN can help in balancing the loss function between the foreground and background pixels.  
Some deep learning methods may use weights of pre-trained networks. A study by \cite{oliveira2018dhsegment} proposed a multi-task document layout analysis approach using Convolution Neural Network (CNN), which adopted transfer learning using ImageNet. 
\cite{yang2017learning} treats the document layout analysis tasks as a pixel-by-pixel classification task. He proposed an end-to-end multi-modal network that contains visual and textual information.

\section{Conclusion}

To empower the document layout analysis research, we present DocBank with 500K high-quality document pages that are built in an automatic way with weak supervision, which enables document layout analysis models using both textual and visual information. To verify the effectiveness of DocBank, we conduct an empirical study with four baseline models, which are BERT, RoBERTa, LayoutLM and Faster R-CNN. Experiment results show that the methods integrating text and layout information is a promising research direction with the help of DocBank. We expect that DocBank will further release the power of other deep learning models in document layout analysis tasks.



\section{Acknowledgments}
This work was supported in part by the National Natural Science Foundation of China (Grant Nos.U1636211, 61672081, 61370126), the Beijing Advanced Innovation Center for Imaging Technology (Grant No.BAICIT-2016001), and the Fund of the State Key Laboratory of Software Development Environment (Grant No.SKLSDE-2019ZX-17).

\bibliographystyle{coling}
\bibliography{coling2020}

\end{document}